\documentclass[10pt,twocolumn,letterpaper]{article}

\usepackage{iccv}
\usepackage{times}
\usepackage{epsfig}
\usepackage{graphicx}
\usepackage{amsmath}
\usepackage{amssymb}

\usepackage{caption}
\usepackage{subcaption}
\usepackage{booktabs}
\usepackage{array,multirow}
\usepackage{xcolor}
\usepackage[british,american]{babel}
\usepackage{blindtext}
\usepackage[inline]{enumitem}
\usepackage{balance}
\usepackage{soul}
\usepackage[export]{adjustbox}

\usepackage[pagebackref=true,breaklinks=true,letterpaper=true,colorlinks,bookmarks=false]{hyperref}
\usepackage[font=small]{subcaption}
\let\oldparagraph\paragraph
\renewcommand{\paragraph}[1]{\vspace{-0.45cm} \oldparagraph{#1}}
\newcommand{\figref}[1]{\mbox{Fig.~\ref{#1}}}

\newcommand{\secref}[1]{\mbox{Sec.~\ref{#1}}}
\renewcommand{\eqref}[1]{\mbox{Eq.~\ref{#1}}}

\newcommand{\retrievalmanifold}{retrieval simplex}
\newcommand{\drn}[1]{\mbox{DRN#1}}

\newcommand*{\affmark}[1][*]{\textsuperscript{#1}}
\newcommand*{\email}[1]{\tt\small{#1}}

\newif\ifedit

\iccvfinalcopy 

\def\iccvPaperID{***}

\ificcvfinal\fi

\begin{document}

\title{Traversing the Continuous Spectrum of Image Retrieval \\with Deep Dynamic Models}

\author{
\begin{tabular}{cccccc}
 \multicolumn{2}{c}{Ziad Al-Halah\affmark[1]} &  \multicolumn{2}{c}{Andreas M. Lehrmann} &  \multicolumn{2}{c}{Leonid Sigal\affmark[2,3]}  \\
 \multicolumn{2}{c}{\email{ziadlhlh@gmail.com}} &  \multicolumn{2}{c}{\email{andreas.lehrmann@gmail.com}} &  \multicolumn{2}{c}{\email{lsigal@cs.ubc.ca}} \\
 \multicolumn{2}{c}{\affmark[1]Karlsruhe Institute of Technology} &  \multicolumn{2}{c}{\affmark[2]Vector Institute} &  \multicolumn{2}{c}{\affmark[3]University of British Columbia} 
\end{tabular}\\
}

\maketitle

\begin{abstract}
\vspace{-0.2cm}
We introduce the first work to tackle the image retrieval problem as a continuous operation.
While the proposed approaches in the literature can be roughly categorized into two main groups: category- and instance-based retrieval, in this work we show that the retrieval task is much richer and more complex.
Image similarity goes beyond this discrete vantage point and spans a continuous spectrum among the classical operating points of category and instance similarity.
However, current retrieval models are static and incapable of exploring this rich structure of the retrieval space since they are trained and evaluated with a single operating point as a target objective.
Hence, we introduce a novel retrieval model that for a given query is capable of producing a dynamic embedding that can target an arbitrary point along the continuous retrieval spectrum.
Our model disentangles the visual signal of a query image into its basic components of categorical and attribute information. 
Furthermore, using a continuous control parameter our model learns to reconstruct a dynamic embedding of the query by mixing these components with different proportions to target a specific point along the retrieval simplex. 
We demonstrate our idea in a comprehensive evaluation of the proposed model and highlight the advantages of our approach against a set of well-established discrete retrieval models.
 
\end{abstract}

\vspace{-0.5cm}
\section{Introduction}\label{sec:introduction}

Image retrieval is a primary task in computer vision and a vital precursor for a wide range of topics in visual search. 
The core element of each retrieval algorithm is a procedure that queries an image database and returns a ranked list of images that are close to the query image.
The ranking is defined with respect to a retrieval objective and a corresponding distance metric.
The retrieval objective can be any underlying property of an image, such as its categories (\eg, car, table, dog)~\cite{gong2013iterative,hwang2012learning} or its visual attributes (\eg, metallic, hairy, soft)~\cite{felix2012weak,siddiquie2011image}.
This objective is expressed during training with a suitable criterion (\eg, a cross entropy loss for categories) to encode the relevant information in a learned, low dimensional feature space (\ie, the embedding space)~\cite{jegou2012negative,perronnin2010large}.
The distance metric is typically an Euclidean distance or a learned metric that captures the pairwise similarity in the respective embedding space.

\begin{figure}[t]
\centering
    \includegraphics[width=0.8\linewidth]{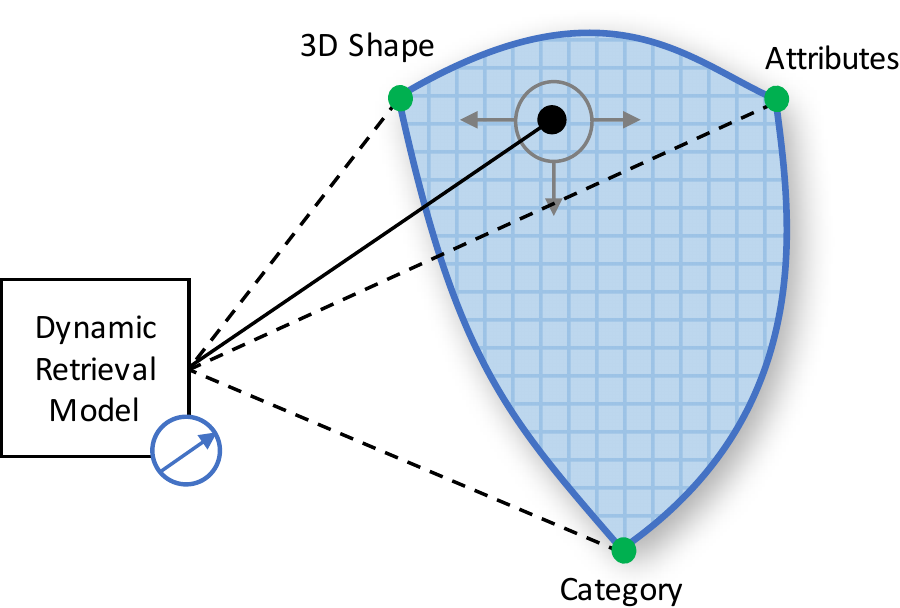}
\caption{Unlike classical \emph{discrete} retrieval models operating at a single point like category- or attribute-based retrieval (green points), we propose to consider the full \emph{continuous} spectrum of the retrieval simplex (blue surface). Our model dynamically constructs a query embedding to target an arbitrary operating point on the simplex (black point).}
\label{fig:intro}
\vspace{-0.4cm}
\end{figure}
 
While this classical approach works well in practice, it is also static and inflexible.
Once the model is trained, we can only retrieve images based on the single retrieval objective chosen at training time, \eg, images with similar attributes or with similar categories. 
However, the space of image-metadata is rich.
Each new type of hand-annotated or machine-inferred data constitutes an independent axis of interest that retrieval models should account for (\eg, objects, captions, and other types of structured data related to the composition of a scene).
A simple way to incorporate such diverse objectives during the retrieval process is to learn a joint embedding with a hyper approach~\cite{Liu_2017_CVPR}.
This is not desirable for the following reasons: 
(1) It reduces the amount of available training data to the point where standard training techniques become infeasible. 
(2) The semantic axes of interest are often orthogonal (\eg, a cat can be white, but not all cats are white and not all white objects are cats), hence augmenting the label space through such a coupling is neither flexible nor scalable.
(3) The contributions of the individual objectives are fixed and unweighted.

Instead, we propose a novel retrieval concept that accounts for retrieving images based on a convex combination of multiple objectives in a continuous manner (\figref{fig:intro}).
Hence, any valid weighting on this simplex of retrieval objectives can be chosen at \emph{test-time}.
We refer to a specific point on this simplex as the~\emph{simplicial retrieval operating point (SROP)}.
In this work, we explore a setting with two retrieval objectives: categories and attributes.
We propose a novel approach that allows targeting a specific SROP at test time.
The resulting model can be viewed as a continuous slider that can retrieve images based on the query image's category, its attributes or any other SROP between those two extremes (\eg, an equal weighting between both).
The representation disentangling between categories and attributes is achieved through parallel memory networks learning corresponding prototypes. 
In particular, we assign a memory for generalization, where we capture categorical prototypes, and another one for specification, where we capture attribute-based prototypes.
Both memories are learned end-to-end based on gated backpropagation of gradients from custom losses, one for each respective objective.
Crucially, both the gates and the losses are SROP-weighted.
At test-time, our model can \emph{dynamically} predict the suitable embedding of an image for a targeted SROP by retrieving and mixing the relevant prototypes learned in both memories.

\paragraph{Contributions}
We make the following contributions in this work: 
(1) we introduce the first work to address image retrieval as a continuous task, namely spanning the continuous spectrum between different operating points; 
(2) we propose a novel \emph{continuous} retrieval module allowing test-time selection of a simplicial retrieval operating point in a smooth and dynamic fashion; 
(3) we introduce and validate novel optimization techniques that are necessary for efficient and effective learning of the retrieval modules; 
(4) we evaluate the advantages of our approach against the classical \emph{discrete} deep retrieval models and demonstrate its effectiveness in a real world application of visual fashion retrieval.
 
\section{Related Work}\label{sec:related_work}

The bulk of image retrieval research can be split into two main groups: instance- and category-based retrieval.
In \emph{instance-based} retrieval we want to retrieve images of the exact same object instance presented in a query (potentially in a different pose or from a different viewing angle).
Early work in this direction focused on matching low-level descriptors~\cite{jegou2010aggregating,nister2006scalable,philbin2007object,sivic2003video}, learning a suitable similarity metric~\cite{jegou2010improving,mikulik2013learning}, or compact representation~\cite{Cao_2018_CVPR,jegou2012negative,perronnin2010large}.
More recently, deep neural networks became predominant in the retrieval literature.
CNNs, in particular, were used as off-the-shelf image encoders in many modern instance retrieval models~\cite{azizpour2015generic,babenko2015aggregating,babenko2014neural,gong2014multiA,Noh_2017_ICCV,paulin2015local,perronnin2015fisher,radenovic2016cnn,razavian2014cnn,razavian2016visual,tolias2015particular}.
Moreover, siamese~\cite{oh2016deep} and triplet networks~\cite{gordo2016deep,hoffer2015deep,wang2014learning,Yu_2018_ECCV} demonstrated impressive performance as they provide an end-to-end framework for both embedding and metric learning.

\emph{Category-based} retrieval methods are on the other end of the retrieval spectrum.
Here, the models target the semantic content of the query image rather than the specific instance.
For example, given a query image of a house on the river bank, we are interested in retrieving images with similar semantic configuration (\ie, \emph{house}+\emph{river}) rather than images of the exact \emph{blue} house with \emph{red} door in the query.
This type of model learns a mapping between visual representations and semantic content, which can be captured by image tags \cite{gong2014multi,gong2013iterative,hwang2012learning,Ranjan_2015_ICCV}, word vector embeddings~\cite{Frome2013,Norouzi2014}, image captions~\cite{Gordo_2017_CVPR}, or semantic attributes~\cite{felix2012weak,siddiquie2011image,Huang_2015_ICCV}.
Most recently, hyper-models \cite{Li_2018_CVPR,Liu_2017_CVPR,veit2017conditional} are proposed to learn an image embedding by jointly optimizing for multiple criteria like category and attribute classification.
However, these models operate at a fixed mixture point of the two retrieval spaces and not dynamically in between.

Differently, this work is the first to phrase the retrieval task as a continuous operation.
Moreover, we propose a novel model that not only can operate at the two extremes of the retrieval spectrum but is also capable of dynamically traversing the simplex in between, generating intermediate retrieval rankings.
Hence, it effectively provides an infinite set of retrieval models in one and can target the desired operating point at test-time using a control parameter.

Our model is based on memory networks~\cite{sukhbaatar2015end,weston2014memory} which have proven useful in many applications, including text-based \cite{bordes2015large,pmlr-v48-kumar16,miller2016key,sukhbaatar2015end,weston2014memory} and vision-based \cite{tapaswi2016movieqa,xiong2016dynamic,xu2016ask} question answering and visual dialogs \cite{das2016visual,seo2017visual}. 
Here, we present a novel type of memory network architecture that learns and stores visual concept prototypes. We propose two types of memories to distinguish between categorical- and instance-based information.
Moreover, we provide key insights on how to improve learning of these concepts in memory using a novel dropout-based optimization approach. 
 
\section{Dynamic Retrieval Along a Simplex}\label{sec:approach}

\begin{figure}[t]
\centering
    \includegraphics[width=\linewidth]{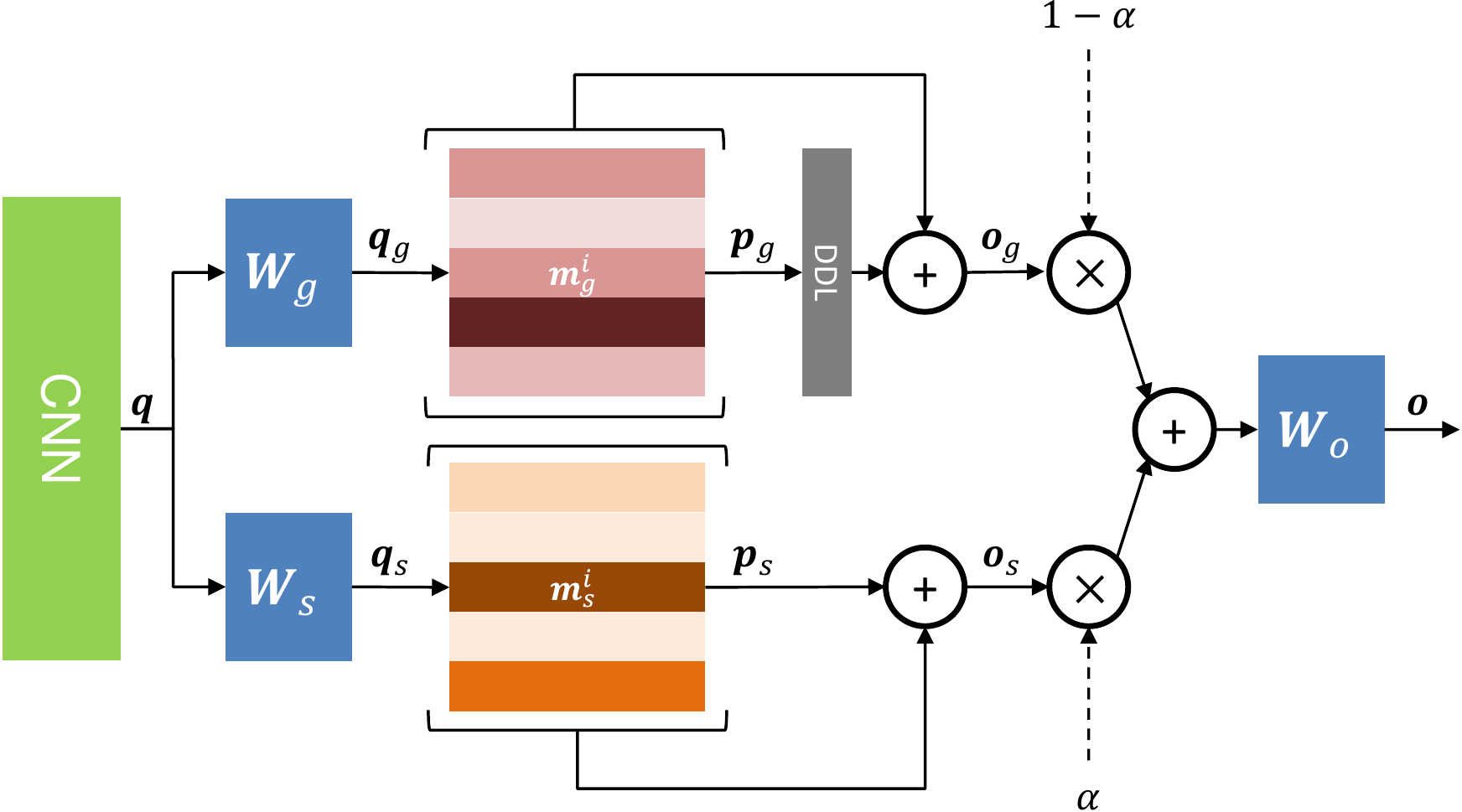}
\caption{\small {\bf Dynamic Retrieval Network.} Two parallel memory modules encode category and attribute prototypes that are fused based on a test-time selection of mixing coefficients on a simplex.}
\label{fig:model_memory}
\vspace{-0.5cm}
\end{figure}

We propose an image retrieval approach that operates dynamically along the retrieval simplex.
We achieve this with a deep model that distills the visual signal from a query image into its basic components (in this case category- and attribute-relevant components) and then blends them to target an arbitrary retrieving operation point. 
Specifically, using a novel memory network architecture (\figref{fig:model_memory}), our model learns various visual prototypes present in the data and stores them in the memory. 
The prototypes can be understood as non-linear sets of basis vectors separately encoding category- and attribute- information.
Given a query image and a control parameter, our model learns to retrieve the relevant prototypes from memory and combines them in appropriate proportions.
Thus, it constructs a custom embedding of the query to target a specific operating point on the retrieval simplex (SROP).

\paragraph{Overview}
Given an image $x$ encoded with $\mathbf{q}=f(x)$ (\eg, a CNN), our proposed dynamic retrieval network (\drn{}) has the following main components:
1) A \emph{query module}, which projects the generic query embedding $\mathbf{q}$ into category- ($\mathbf{q}_g$) and attribute-specific ($\mathbf{q}_s$) representations to address our parallel memory module;
2) A \emph{memory module}, which learns visual concept prototypes in a generalization ($\mathbf{M}_g$) and a specification ($\mathbf{M}_s$) memory. 
Both $\mathbf{q}_{\bullet}$ and $\mathbf{M}_{\bullet}$ are combined to form a category ($\mathbf{o}_g$) and attribute ($\mathbf{o}_s$) representation for the query image.
Finally, 3) An \emph{output module}, which mixes the constructed representations with different proportions given a \emph{specificity} control parameter $\alpha$.
The query embedding derived by the output module is then compared (\eg, using Euclidean distance) to similarly computed embeddings of each image in the dataset to arrive at the ranked list of retrieved images.
Next, we provide a detailed description of our \drn{} and its components.

\vspace{-0.17cm}
\subsection{Query module}
\vspace{-0.17cm}
Given an image $x$ encoded with a deep network, ${\bf q}=f(x)$, we query each memory module using $\mathbf{q}_s$ and $\mathbf{q}_g$.
These two queries are learned using two different projection layers $f_s(\cdot)$ and $f_g(\cdot)$ to adapt to the different representations in $\mathbf{M}_g$ and $\mathbf{M}_s$:
\begin{align}
\begin{split}
	\mathbf{q}_g &= f_g(x) = \mathbf{W}_gf(x)+b_g,\\
	\mathbf{q}_s &= f_s(x) = \mathbf{W}_sf(x)+b_s,	
\end{split}
\end{align}
where $\mathbf{W}_{\bullet}$ and $b_\bullet$ are the weights and bias for each layer.

 \vspace{-0.17cm}
\subsection{Memory Modules}
\vspace{-0.17cm}
To operate dynamically at an arbitrary SROP, we need to control the information embedded in the query representation.
Hence, we propose to factorize the representation of an input $x$ in terms of a category- and attribute-based representation.
While the category-based representation captures the shared information between $x$ and all samples of the same category as $x$, the attribute-based representation captures the information that distinguishes $x$ from the rest (\ie, its visual attributes).
By separating the two signals, we control how to mix these two representations to construct a new embedding of $x$ that targets the desired SROP on the retrieval simplex.
Hence, our memory module is made up of two parallel units:
a) The \emph{generalization memory} ($\mathbf{M}_g$), that learns category-based prototypes;
and b) The \emph{specification memory} ($\mathbf{M}_s$), that learns attribute-based prototypes.

\paragraph{Generalization Memory} 
Here, we would like to learn concept prototypes that capture information shared among all samples of the same category.
Based on the intuitive assumption that samples usually belong to a single (or a few) base categories, we can use a softmax layer to attend to the memory cells of $\mathbf{M}_g$.
Given the sparse properties of a softmax layer, this allows us to learn a discriminative category-based representations and attend to the most suitable one to construct the input embedding.
That is, given a query $\mathbf{q}_g$, the category-based embedding of $\mathbf{q}_g$ is constructed as:
\begin{align}
\begin{split}
	p_g^i &= \mathrm{softmax}(\mathbf{q}_g^\top \mathbf{m}_g^i),\\
	\mathbf{o}_g &= \sum_i^{N_g}{p_g^i \mathbf{m}_g^i},
\end{split}
\end{align}
where $p_g^i$ is the attention over memory cell $\mathbf{m}_g^i$, $N_g$ is the total number of generalization memory cells, and $\mathbf{o}_g$ is the output of the generalization memory module $\mathbf{M}_g$.

\paragraph{Dropout Decay Layer (DDL)}
While the sparsity of the softmax is beneficial to learn discriminative prototypes, it may result in stagnation during the optimization process since the backpropagated error signals get channeled to a few memory cells. 
To counter these optimization difficulties caused by the softmax layer, we propose a \emph{dropout decay layer} (DDL) over the attentions $p_g$.
The DDL is a dropout layer with an initial dropout probability $p_d$ and a decay factor $\gamma_d$.
Starting with an initially high dropout rate $p_d$, the DDL pushes our model to activate and initialize many cells in $\mathbf{M}_g$ rather than relaying on just a few. 
As the training progresses, the dropout probability gradually gets dampened by $\gamma_d$, enabling the model to learn a more discriminative representation in $\mathbf{M}_g$. 

\paragraph{Specification Memory}  
While each sample typically belongs to a single base category, it usually exhibits a set of multiple visual attributes that are instance specific and distinguish this instance from others within the same category.
Hence, we use a sigmoid layer to attend to the memory cells of $\mathbf{M}_s$, which allows us to select and compose multiple attributes.
Given a query ${\bf q}_s$, we construct the attribute-based representation $\mathbf{o}_s$ as follows: 
\begin{align}
\begin{split}
	p_s^j &= \mathrm{sigmoid}(\mathbf{q}_s^\top \mathbf{m}_s^j),\\
	\mathbf{o}_s &= \frac{1}{\sum_jp^j_s}\sum_j^{N_s}{p_s^j \mathbf{m}_s^j},
\end{split}
\end{align}
where $p_s^j$ is the attention over memory cell $\mathbf{m}_s^j$, $N_s$ is the number of cells in $\mathbf{M}_s$, and $\mathbf{o}_s$ is the output of the specification memory module.
Unlike softmax, the sigmoid layer does not produce a sparse attention over $\mathbf{M}_s$ and hence we do not need a special activation mechanism as in $\mathbf{M}_g$.

\subsection{Output Module}
Given the output representation of the two memory modules, we construct the representation of input sample $x$ as a weighted linear combination of $\mathbf{o}_s$ and $\mathbf{o}_g$:
\begin{equation}
	\mathbf{o} = f_o(\mathbf{o}_s,\mathbf{o}_g) = \mathbf{W}_o(\alpha\cdot\mathbf{o}_s + (1-\alpha)\cdot\mathbf{o}_g) + b_o,
\end{equation}
where $f_o(\cdot)$ is an embedding of the linear combination of the memory modules' outputs.
The specificity control $\alpha \in [0,1]$ weights the contribution of each memory module in the final representation of $x$.
As $\alpha$ approaches zero, the category-based information of $x$ is emphasized in $\mathbf{o}$.
By increasing $\alpha$, we incorporate more instance-specific information (with attributes as proxy) from $x$ into $\mathbf{o}$.

\subsection{Dynamic Learning Objective}\label{app:multitask}
To learn the representations in the memory cells of $\mathbf{M}_g$ and $\mathbf{M}_s$, we need a learning objective that can distill the error signal dynamically into category- and attribute-based signals with proportions similar to the targeted SROP of the constructed embedding. 
We achieve this using multitask learning, \ie, we jointly optimize for two criteria to capture both signals. 
Additionally, both criteria are weighted by the specificity parameter $\alpha$ that controls the contributions of the backpropagated error signals to the memory modules.
We consider two options:

\paragraph{1) Classification-based \drn{} (\drn{-C})}
This model is optimized jointly for category and attribute classification:
\begin{equation}\label{eq:cls_loss}
	\mathcal{L}(x) = (1-\alpha)\mathcal{L}_{cls} (x) + \alpha \mathcal{L}_{att} (x) + \beta \mathcal{L}_{reg}(\theta)
\end{equation}
where $\mathcal{L}_{cls}$ and $\mathcal{L}_{att}$ are cross entropy losses for the categories and attributes, respectively, and $\mathcal{L}_{reg}=||\theta||^2_2$ is a regularization loss over the model parameters $\theta$.

\vspace{0.1cm}
\paragraph{2) Similarity-based \drn{} (\drn{-S})} 
This model optimizes jointly for category classification and pairwise instance similarity: 
\begin{align}
\begin{split}
	\mathcal{L}(x_1,x_2) = &(1-\alpha)(\mathcal{L}_{cls}(x_1)+\mathcal{L}_{cls}(x_2)) +\\ 
	&\alpha \mathcal{L}_{sim} (x_1,x_2)+ \beta \mathcal{L}_{reg}(\theta),
\end{split}
\end{align}
where $\mathcal{L}_{cls}$ and $\mathcal{L}_{reg}$ are defined as above and
$\mathcal{L}_{sim}$ is a margin-based contrastive loss~\cite{hadsell2006dimensionality}. 

\drn{-C} and \drn{-S} have different characteristics:
Since \drn{-C} is optimized with cross entropy losses, we expect a strong discriminative error signal from object and attribute losses, which will help in learning more discriminative prototypes.
On the other hand, \drn{-S} leverages a contrastive loss that captures the generic pairwise similarity of samples in the attribute embedding space.
Experimentally, we find that this lends itself well to cases where it is desirable to maintain the category as $\alpha$ changes.
Intuitively, \drn{-S} can be considered to be more generic than \drn{-C}, since arbitrary similarity metrics can be used in this formulation (\eg, caption similarity) to define the extrema of the retrieval simplex and, consequently, the SROPs that we wish to traverse.

\paragraph{Sampling $\alpha$} For each training image $x$, we sample $\alpha$ randomly from $[0,1]$ using a uniform distribution. 
Note that $\alpha$ not only controls the mixing of category- and attribute-based prototypes, but also controls the error signal coming from the category- and attribute-based losses.
$\alpha$ acts as a gating layer which controls the flow of information to each memory module and allows us to distill the backpropagated error signal during training. 
At test time, we can select $\alpha$ freely to control the SROP, which can be anywhere between pure category-prototypes and pure attribute-based ones.

\section{Evaluation}\label{sec:evaluation}

Our experiments evaluate our model's ability to operate along the \retrievalmanifold.
We first conduct a thorough evaluation of the modules proposed in~\secref{sec:approach} by validating our design choices and their impact on the performance and the learned concept prototypes (\secref{sec:eval:memory}). 
Next, we evaluate both \drn{-C} and \drn{-S} in the proposed retrieval task and highlight their distinct properties.
A comparison to popular baselines demonstrates the advantages of our approach (\secref{sec:comparison}). 
We conclude with the analysis of our model in a fashion retrieval application (\secref{sec:fashion}).

\paragraph{MNIST Attributes}
We evaluate our model on the MNSIT Attribute dataset, which is driven from the MNIST~\cite{lecun1998gradient} and MNIST Dialog~\cite{seo2017visual} datasets.
In particular, each digit image of the $10$ classes is augmented with $12$ binary visual attributes: $5$ foreground colors of the digit, $5$ background colors, and $2$ style-related attributes (\emph{stroke}, \emph{flat}). 
In total, we have $20,000$ images for training, $5,000$ for validation, and $5,000$ for testing (see \figref{fig:alpha_qual_cls_sim}). 

\paragraph{Model Architecture}
In order to guarantee a fair comparison, we use the same core CNN (green box in~\figref{fig:model_memory}) in all our experiments: the first $2$ layers consist of convolutions with kernel size $5$ and $20$/$50$ feature maps, respectively. 
Both layers are followed by a max-pooling layer with stride $2$. 
The core CNN ends with $2$ fully-connected layers with ReLU activations of size $500$. 
We combine this core CNN with the modules illustrated in \figref{fig:model_memory} to obtain our full dynamic model and with various classifiers for our baseline comparisons. 
We train all models using Adam~\cite{Kingma2015} with an initial learning rate of $10^{-4}$.

\begin{figure*}[!t]
\centering
\begin{subfigure}[!b]{0.27\linewidth}
    \centering
    \includegraphics[width=0.7\linewidth]{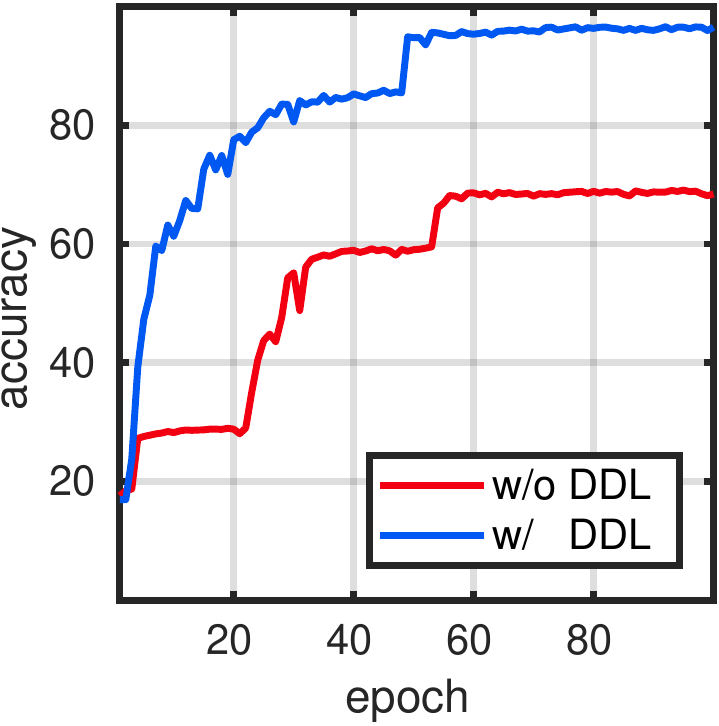}
    \caption{Performance comparison.}
    \label{fig:mem_dropout_v3_a}
\end{subfigure}
\begin{subfigure}[!b]{0.27\linewidth}
    \centering
    \includegraphics[width=0.9\linewidth]{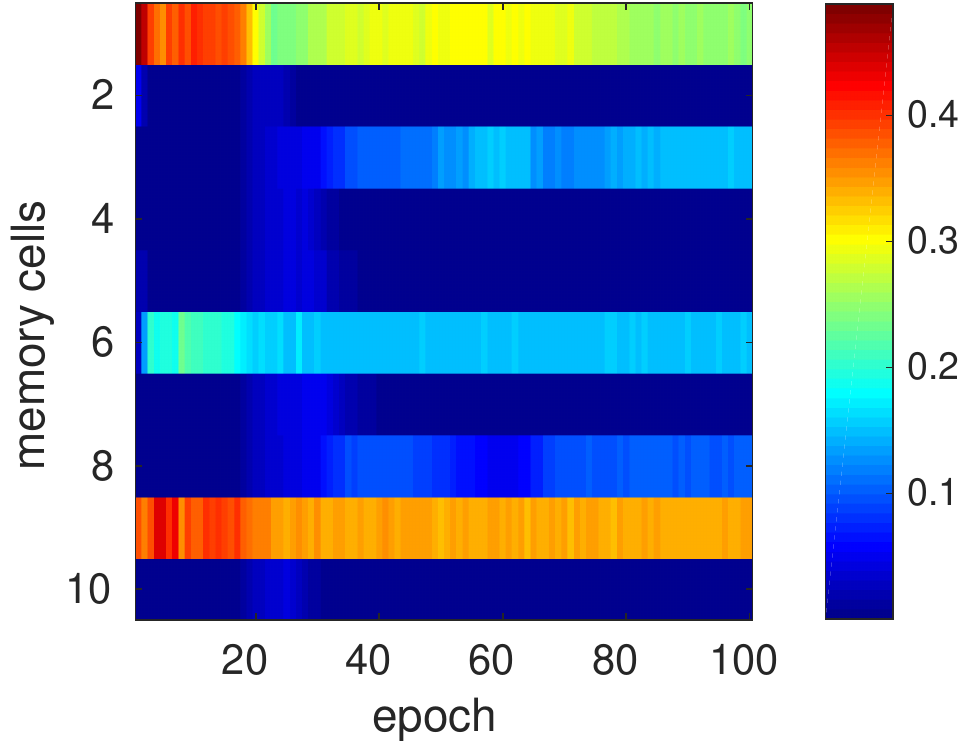}
    \caption{Activation history without DLL.}
    \label{fig:mem_dropout_v3_b}
\end{subfigure}
\begin{subfigure}[!b]{0.27\linewidth}
    \centering
    \includegraphics[width=0.9\linewidth]{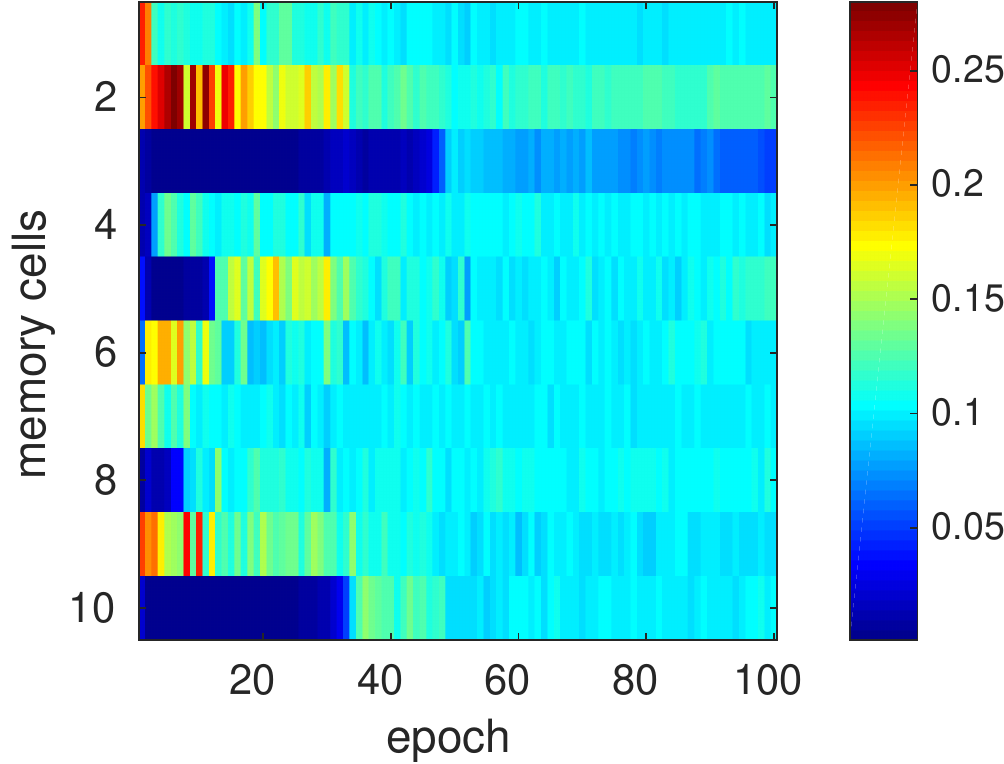}
    \caption{Activation history with DLL.}
    \label{fig:mem_dropout_v3_c}
\end{subfigure}
\caption{\small {\bf Analysis of the dropout decay layer (DDL).} 
{\bf(a)} DDL leads to significant improvements in both performance and convergence rate. 
{\bf(b)} Without DLL, the model tends to rely on a few learned prototypes for extended periods of time, resulting in performance stagnation. 
{\bf(c)} With DLL, the model is incentivized to activate additional memory cells early in the training process.}
\label{fig:mem_dropout_v3}
\vspace{-0.4cm}
\end{figure*} 
\begin{figure*}
\centering
\begin{minipage}[b][][b]{0.48\textwidth}
	\centering
	\includegraphics[width=0.49\linewidth]{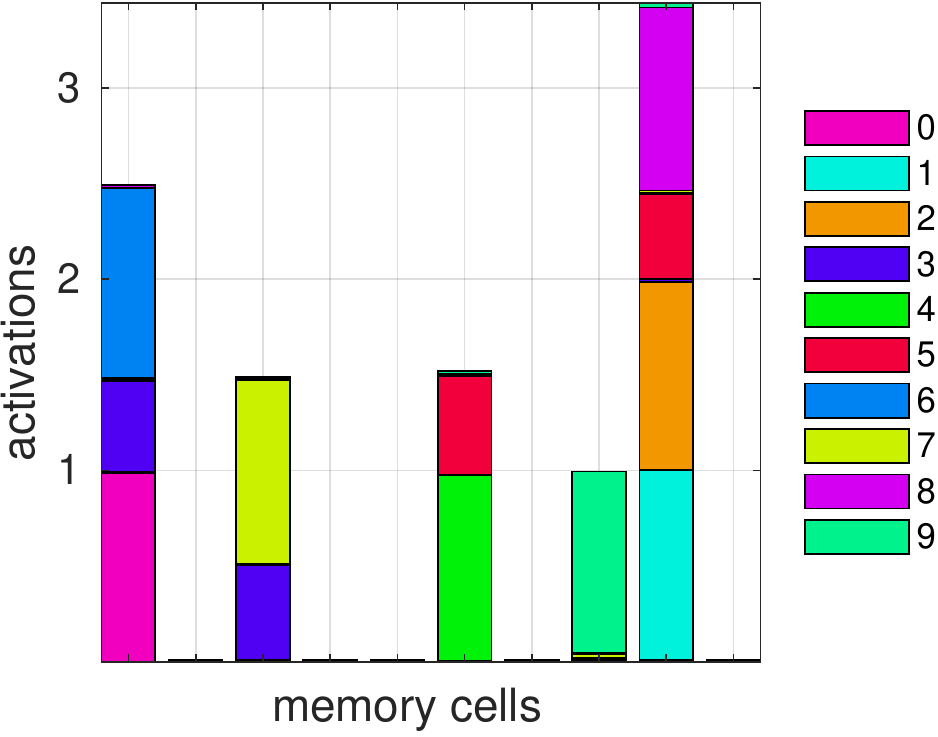}
	\includegraphics[width=0.49\linewidth]{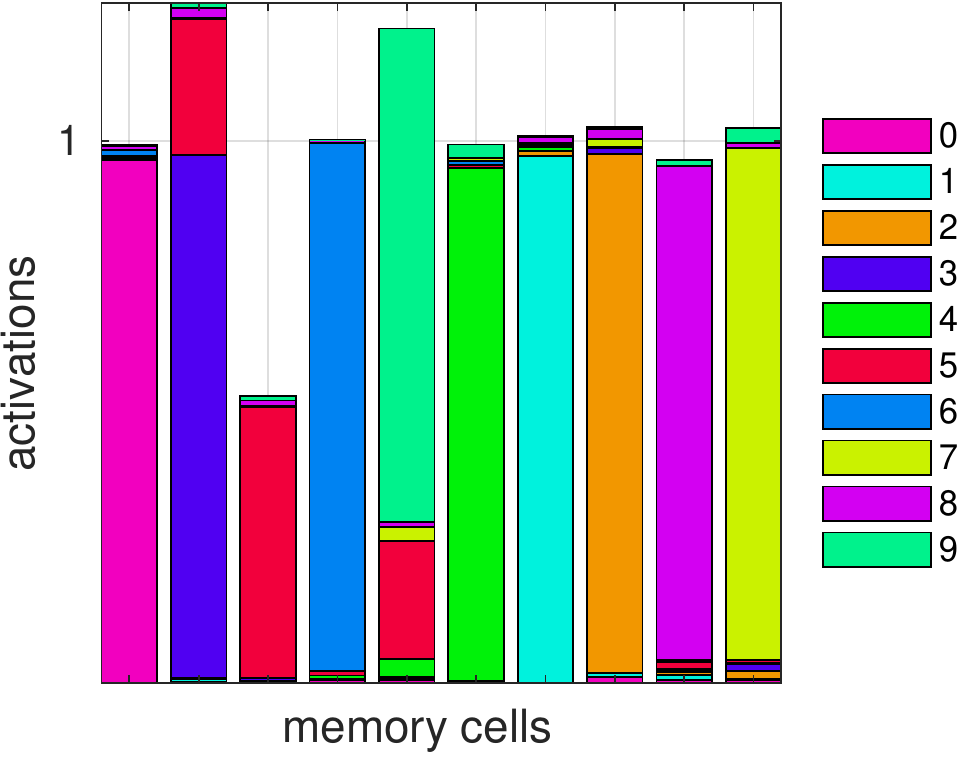}
	\captionof{figure}{\small {\bf Prototypes learned in memory.} 
	Visualization of cell activation {\bf(left)} without dropout decay layer and {\bf(right)} with dropout decay layer. 
	Colors indicate utilization of cells by a particular digit.}
	\label{fig:mem_dropout_activations}
\end{minipage}\hspace{0.5cm}
\begin{minipage}[b][][b]{0.48\textwidth}
	\centering
	\includegraphics[width=0.49\linewidth]{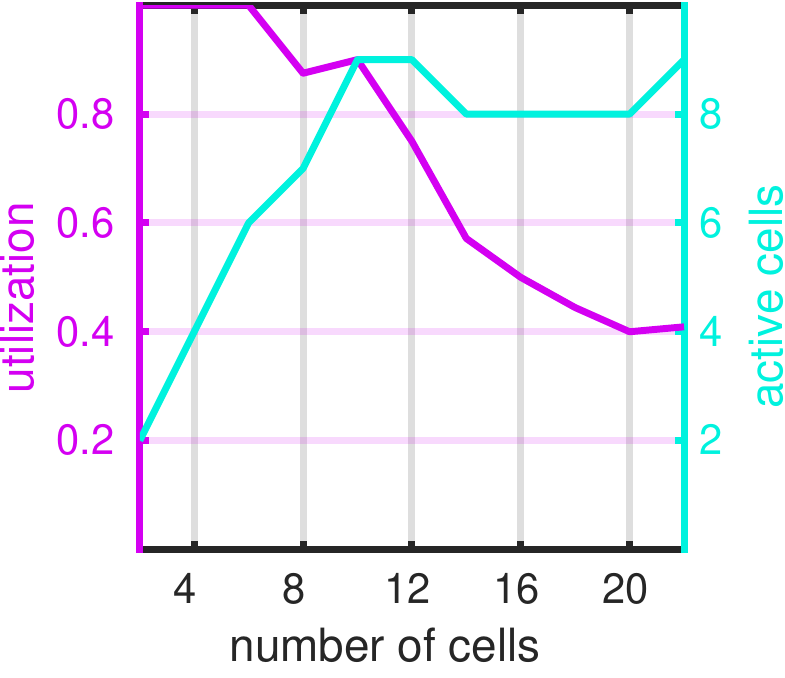}
	\includegraphics[width=0.42\linewidth]{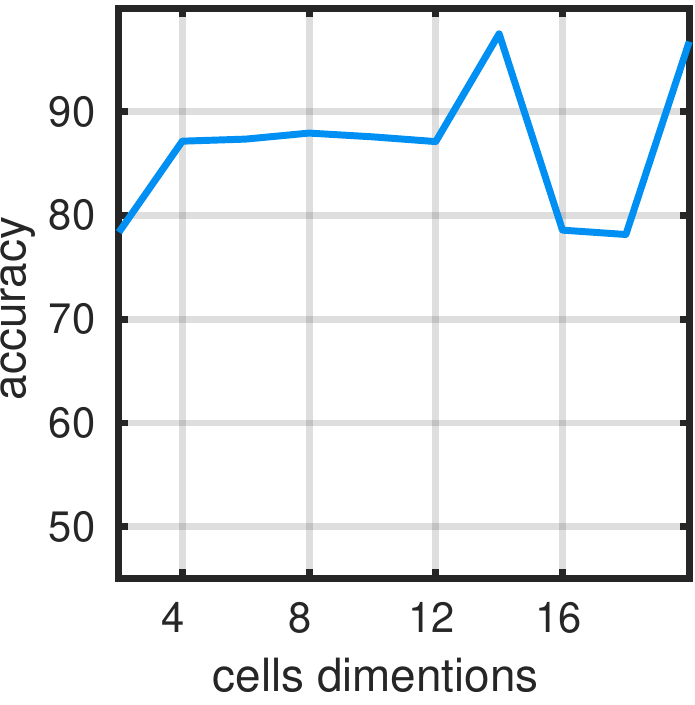}
	\captionof{figure}{\small {\bf Effect of number and dimensionality of cells in $\mathbf{M}_g$.} 
	{\bf(left)} Memory utilization across number of cells. 
	{\bf(right)} Model performance across cell dimensionality.}
	\label{fig:mem_util_dim}
\end{minipage}
\end{figure*} \subsection{Learning Visual Concept Prototypes}\label{sec:eval:memory}
We start by analyzing the properties of our memory modules. 
Hence, we train a CNN augmented with our generalization memory $\mathbf{M}_g$ and followed by a softmax layer for category classification and use a cross-entropy loss.
We set the number of cells in $\mathbf{M}_g$ to $N=10$ and the initial dropout probability to $p_d=0.9$, with a decay of $\gamma_d=10^{-5}$.

\paragraph{Dropout Decay Layer (DDL)}
\figref{fig:mem_dropout_v3_a} shows the performance of our model with and without the proposed DDL. 
The model without DDL suffers from long periods of stagnation in performance (red curve). 
To better understand this phenomenon, we track, during learning, the activation history of the memory cells across the entire training dataset:
examining~\figref{fig:mem_dropout_v3_b}, we notice that overcoming these stages of stagnation actually corresponds to the activation of a new memory cell, \ie, a new category prototype. 
DDL significantly improves the performance of the model by pushing it to activate multiple memory cells early in the training process rather than relying on a few initial prototypes (\figref{fig:mem_dropout_v3_c}). 
Moreover, the early epochs of training are now characterized by high activations of multiple cells, which gradually get dampened in the later stages. 
In summary, DDL not only counteracts the stagnation in training but also results in faster convergence and better overall performance.

\begin{figure}
\centering
\begin{subfigure}[!b]{0.49\linewidth}
    \includegraphics[width=0.8\linewidth]{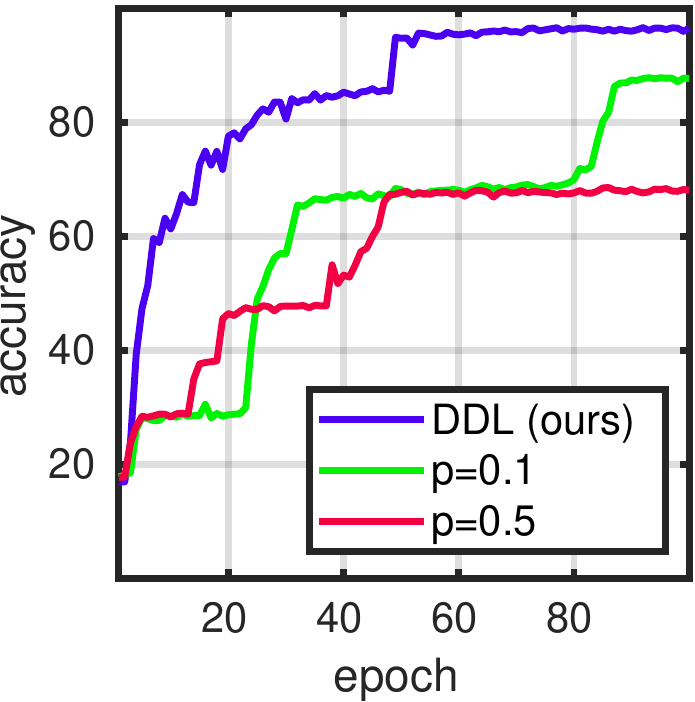}
    \caption{Top-1}
    \label{fig:const_dropout_a}
\end{subfigure}
\begin{subfigure}[!b]{0.49\linewidth}
    \includegraphics[width=\linewidth]{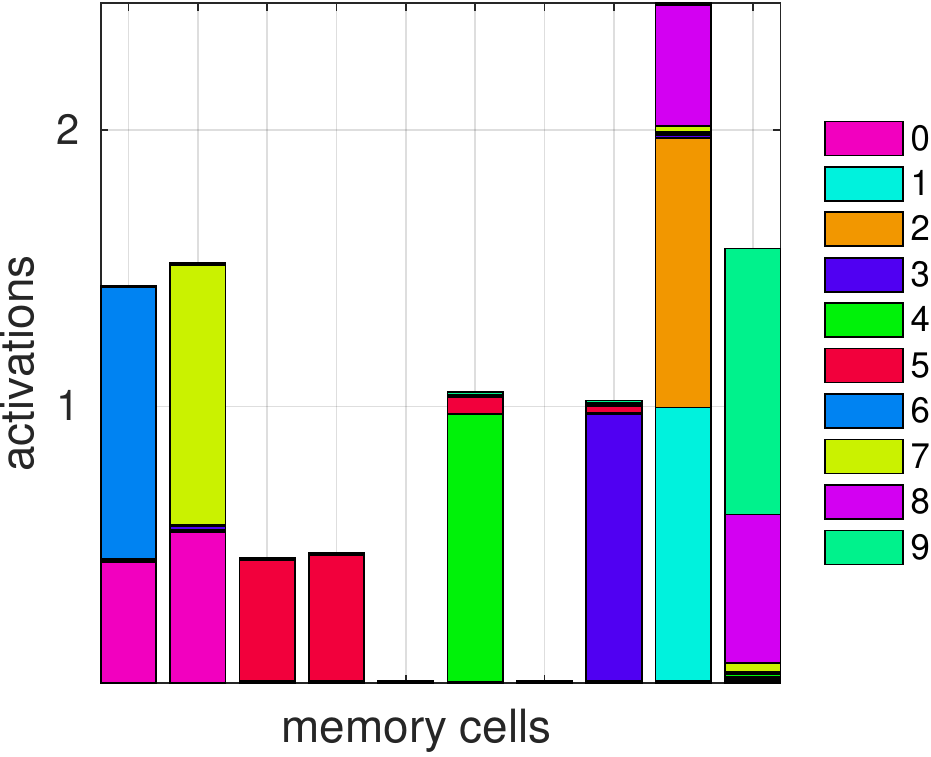}
    \caption{$p = 0.1$}
    \label{fig:const_dropout_d}
\end{subfigure}
\caption{\small {\bf Comparison to constant dropout.} 
{\bf(a)} DDL performs significantly better than a network with constant dropout. 
{\bf(b)} Constant dropout with $p=0.1$ leads to mixing artifacts similar to a network without DDL (cf.~\figref{fig:mem_dropout_activations}(left)).}
\label{fig:const_dropout}
\vspace{-0.4cm}
\end{figure}
 \paragraph{Memory Semantics}
To gauge what the model actually captures in memory, we validate it on test data by accumulating a histogram of each cell's activations over the categories: \figref{fig:mem_dropout_activations}(right) shows the average activations per category for each cell.
We see that our model learns a clean prototype in each cell for each of the categories. 
In comparison to a model without DDL~(\figref{fig:mem_dropout_activations}(left)), the learned representations in memory are substantially more discriminative.

\paragraph{Constant Dropout} 
Interestingly, DDL is also more effective than standard dropout with fixed values $p=0.1$ and $p=0.5$~(\ie~without decay): in addition to improved performance~(\figref{fig:const_dropout_a}), we see in \figref{fig:const_dropout_d} that such a fixed dropout probability leads to learning mixed category prototypes in the memory cells. 
For some categories, their prototypes are split across multiple cells, while other cells capture a generic prototype that represents multiple categories. 
This is expected: due to the constant dropout during training, multiple cells are forced to capture similar category prototypes in order to cover the information loss encountered from dropping out some of the cells.

\paragraph{Memory Configuration}
Our memory module is parametrized by the number of cells $N$ and their dimensionality $d$.
We analyze the impact of these hyperparameters on memory utilization and model performance in the following experiments:
\begin{itemize}[label={\textbullet},topsep=1pt, leftmargin=*,itemindent=8pt,labelsep=4pt]
\item {\em Number of Cells}:
	We vary the number of cells in the memory and check how many of them are utilized. 
	We measure utilization as the percentage of cells activated for at least $5\%$ of the samples. 
	\figref{fig:mem_util_dim}(left) shows that utilization steadily drops beyond $10$ cells while the number of active cell remains stable at $8$ or $9$, which is close to the number of categories in the dataset. 
	We conclude that our model learns a sparse representation of the categories and uses the memory in a way that is consistent with the visual concepts present in the data. 
	In other words, the model is not sensitive to this parameter and can effectively choose an appropriate number of needed active cells.
\item {\em Dimensionality}: 
	We analyze the impact of cell dimensionality on the final performance of the model.
	We set our memory module to $N=10$ cells and vary the dimensionality $d$.
	\figref{fig:mem_util_dim}(right) shows the change in classification accuracy along $d$.
	Our model exhibits good robustness with regard to memory capacity and reaches $89\%$ accuracy even with a low-dimensional representation.
\end{itemize}

\noindent
While we analyzed the performance of the generalization memory, the analysis of specification memory looks similar, despite the denser activation patterns caused by the sigmoid (vs. softmax) attention for cell referencing.

\subsection{From Category- to Attribute-based Retrieval}\label{sec:comparison}
So far, we have evaluated our memory module and the impact of its configuration on utilization and accuracy. 
Next, we examine the performance and properties of our dynamic models along the \retrievalmanifold.

\begin{figure*}[!t]
\centering
\begin{subfigure}[!b]{0.47\linewidth}
    \centering
    \includegraphics[width=0.49\linewidth]{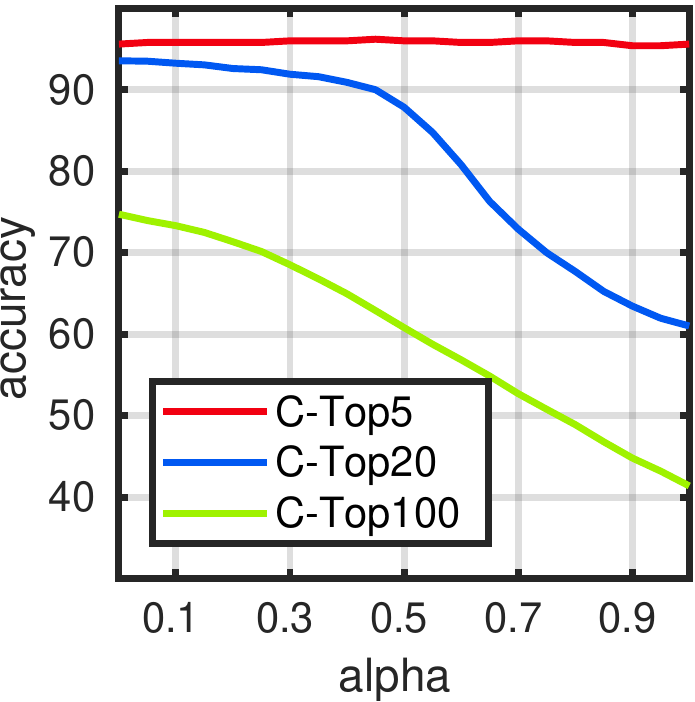}
    \includegraphics[width=0.49\linewidth]{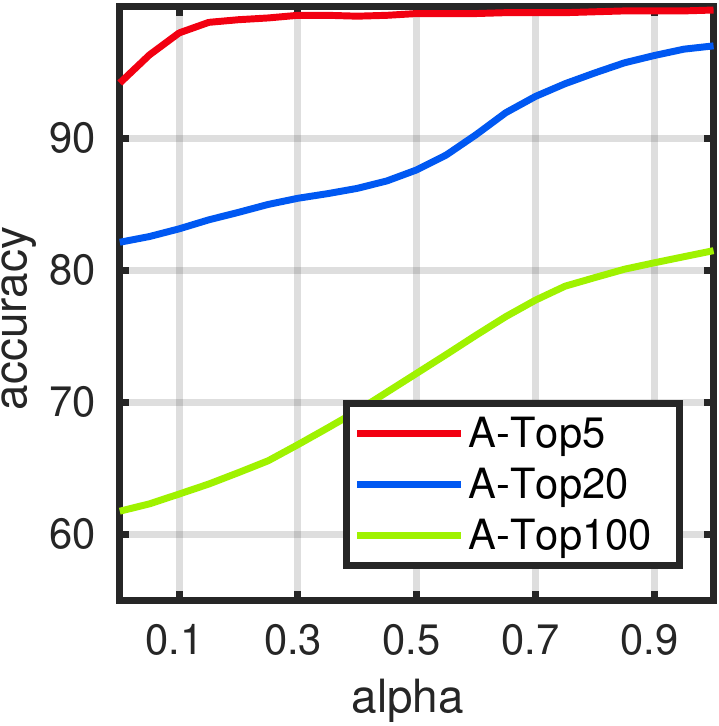}
    \caption{Classification-based model (\drn{-C}).}
    \label{fig:cbmrn}
\end{subfigure}\qquad
\begin{subfigure}[!b]{0.47\linewidth}
    \centering
    \includegraphics[width=0.49\linewidth]{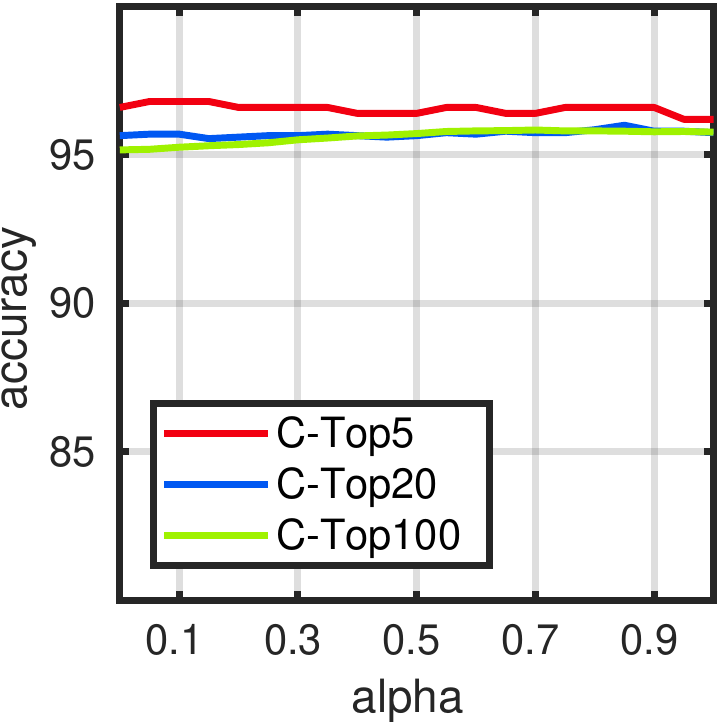}
    \includegraphics[width=0.49\linewidth]{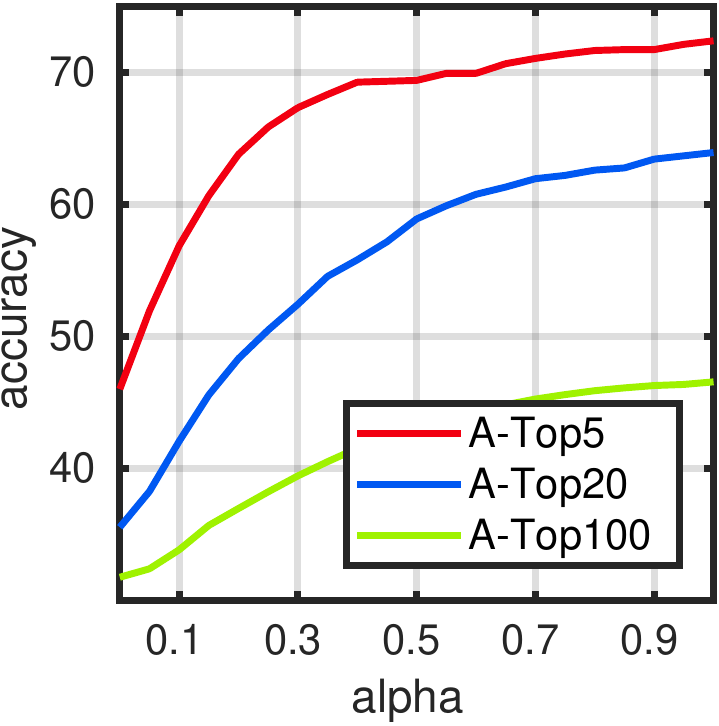}
    \caption{Similarity-based model (\drn{-S}).}
    \label{fig:sbmrn}
\end{subfigure}
\caption{\small {\bf Traversing the \retrievalmanifold.} 
We show the performance of our {\bf(a)} \drn{-C} and {\bf(b)} \drn{-S} as $\alpha$ increases from $0$ to $1$.}
\label{fig:alpha_acc_cls_sim}
\end{figure*} 
\begin{figure*}[!t]
\centering
\begin{subfigure}[!b]{0.5\linewidth}
    \centering
    \includegraphics[width=0.95\linewidth]{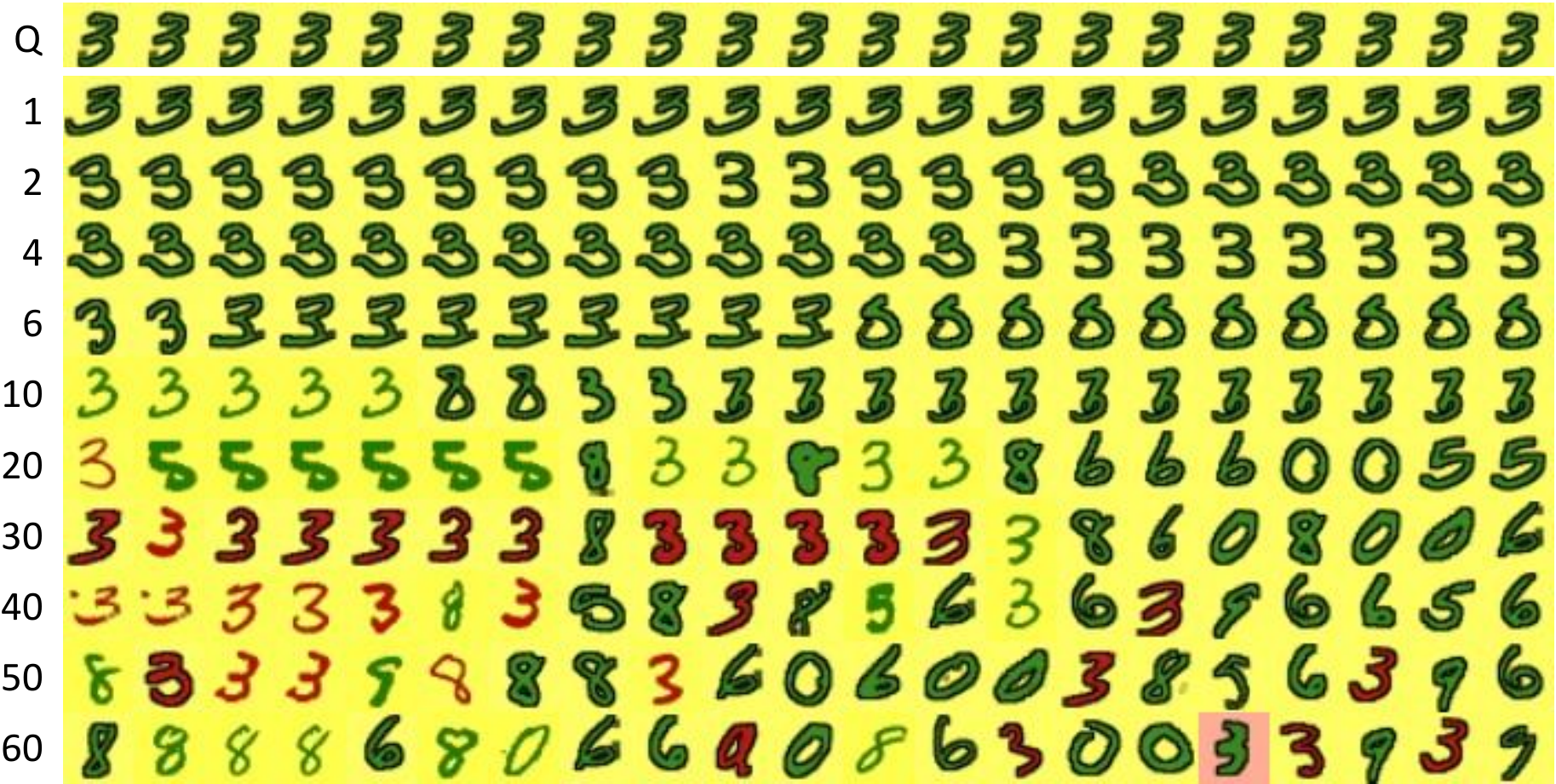}
    \caption{Classification-based retrieval (\drn{-C}).}
    \label{fig:cbmrnQual}
\end{subfigure}\begin{subfigure}[!b]{0.5\linewidth}
    \centering
    \includegraphics[width=0.95\linewidth]{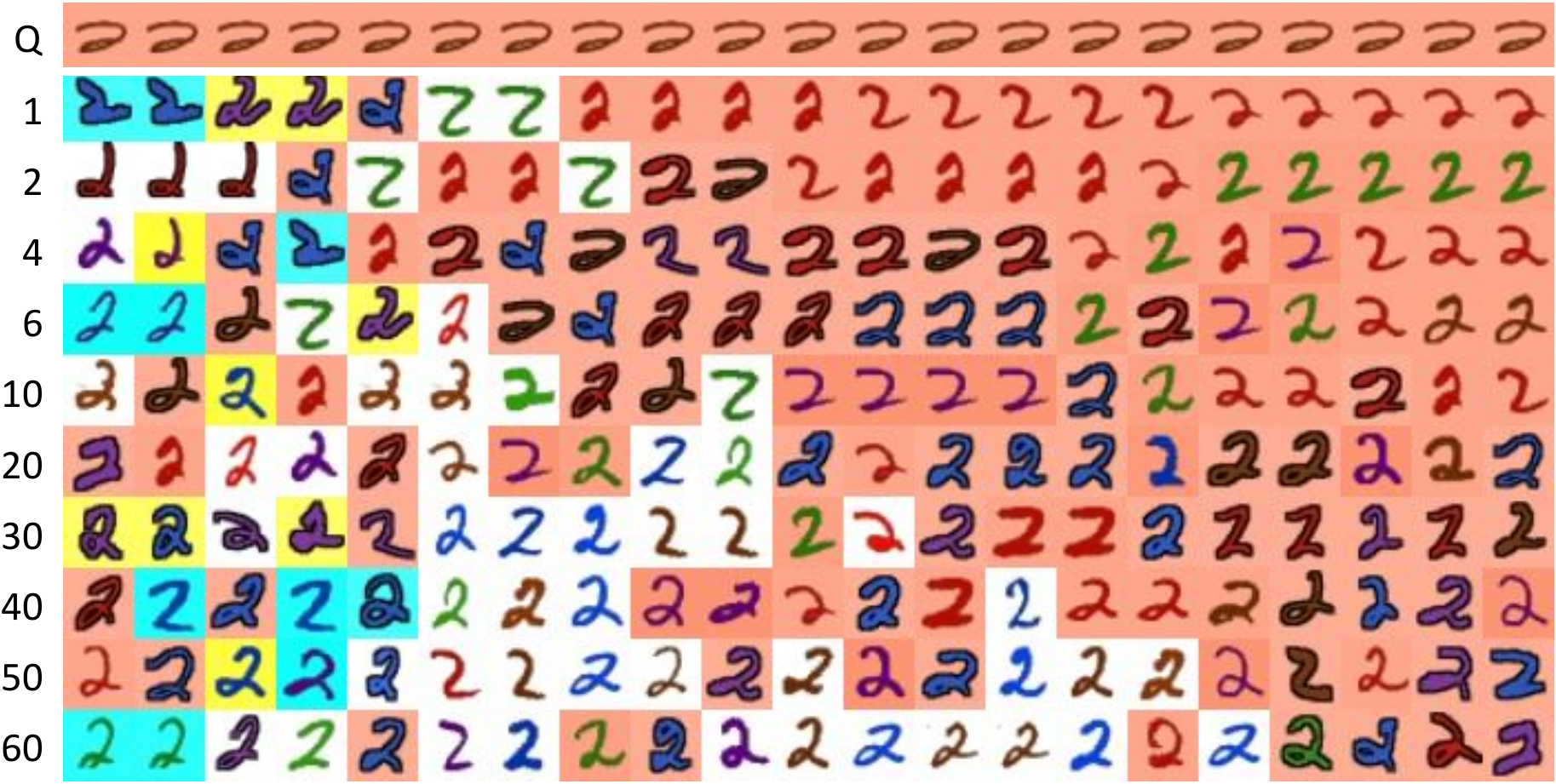}
    \caption{Similarity-based retrieval (\drn{-S}).}
    \label{fig:sbmrnQual}
\end{subfigure}
\caption{\small {\bf Qualitative results.} 
The query (Q) is shown in the first row, the results with indicated rank $K$ in the subsequent rows. 
Each colum corresponds to the result for a particular value of $\alpha$, from $0$ (first column; focused on classification) to $1$ (last column; focused on attributes).
({\bf a}) The \drn{-C} model retrieves the correct class for $\alpha=0$ and smoothly changes to the correct attributes for $\alpha=1$ (neglecting class membership). 
({\bf b}) The \drn{-S} model keeps the correct class membership while increases attribute similarity as $\alpha$ approaches $1$.}
\label{fig:alpha_qual_cls_sim}
\end{figure*} 
We train our full \drn{} models with both generalization and specification memories using the dynamic learning objectives.
The CNN encoder in this experiment is identical to the one we used in \secref{sec:eval:memory}. 
However, in this experiment, we train our model for $2$ objectives that capture category- and attribute-based information, as explained in \secref{app:multitask}.
While \drn{-C} is trained for a joint objective of category and attribute classification, the \drn{-S} model is trained for category classification and pairwise similarity, \ie, a sample pair is deemed similar if they share the same attributes.
During training, we sample $\alpha$ uniformly in the range $[0,1]$ to enable the models to learn how to mix the visual concept prototypes at different operating points.

We measure the performance of category-based retrieval using C-TopK accuracy, which measures the percentage of the top $K$ similar samples with matching category label to the query $q$.
We measure the attribute-based performance using the A-TopK accuracy, which measures the percentage of matching attributes in the top $K$ similar samples to $q$.
At test time, we randomly sample $10$ queries from each category and rank the rest of the test data with respect to the query using Euclidean distance as the similarity metric.
The samples are embedded by the dynamic output of our \drn{} while gradually changing $\alpha$ from $0$ to $1$ to target the different SROPs.

\figref{fig:alpha_acc_cls_sim} shows the retrieval performance of the proposed \drn{-C} and \drn{-S}.
As $\alpha$ goes from $0$ to $1$, category-based accuracy (C-Top) decreases and attribute-based accuracy (A-Top) increases in the classification-based model \drn{-C}, as expected (\figref{fig:cbmrn}). 
The similarity-based model \drn{-S}, on the other hand, shows a stable, high performance for the category-based retrieval with varying $\alpha$ and shares similar properties of the attribute-based performance with the previous model, albeit with larger dynamic range (\figref{fig:sbmrn}).
In summary, \drn{-C} helps us to cross the category boundary of the query to retrieve instances with similar attributes but from different categories.
\drn{-S} model traverses the manifold within the category boundaries and retrieves instances from the same category as the query with controllable attribute similarity.
These conclusions are reinforced by our qualitative results, which we discuss next.

\paragraph{Qualitative Results}
\figref{fig:alpha_qual_cls_sim} shows qualitative results of both models: the first row (Q) is the query and each column displays the top $K$ retrieved instances as alpha goes from $0$ (left-most column) to $1$ (right-most column).
In case of the \drn{-C} model (\figref{fig:cbmrnQual}), we see that the left-most column contains samples with the correct category of the query but with diverse attributes. 
As we move toward $\alpha=1$, we retrieve instances with more matching attributes but also from increasingly more diverse categories. With the \drn{-S} model (\figref{fig:sbmrnQual}), we can traverse from $\alpha=0$ to $\alpha=1$ while keeping the categorical information of the query, with increasing similarity of instance-specific information.
We note that both of these behaviors may be desirable, depending on the application. 
For example, for a generic image retrieval, control over {\em specificity} of results may be desired, as is achieved with the \drn{-S} model; however, for exploring fashion apparel (\secref{sec:fashion}) with a given style, the \drn{-C} model may be more appropriate.

\begin{figure}[!t]
\centering
\begin{subfigure}[!b]{0.45\linewidth}
    \centering
    \includegraphics[width=\linewidth]{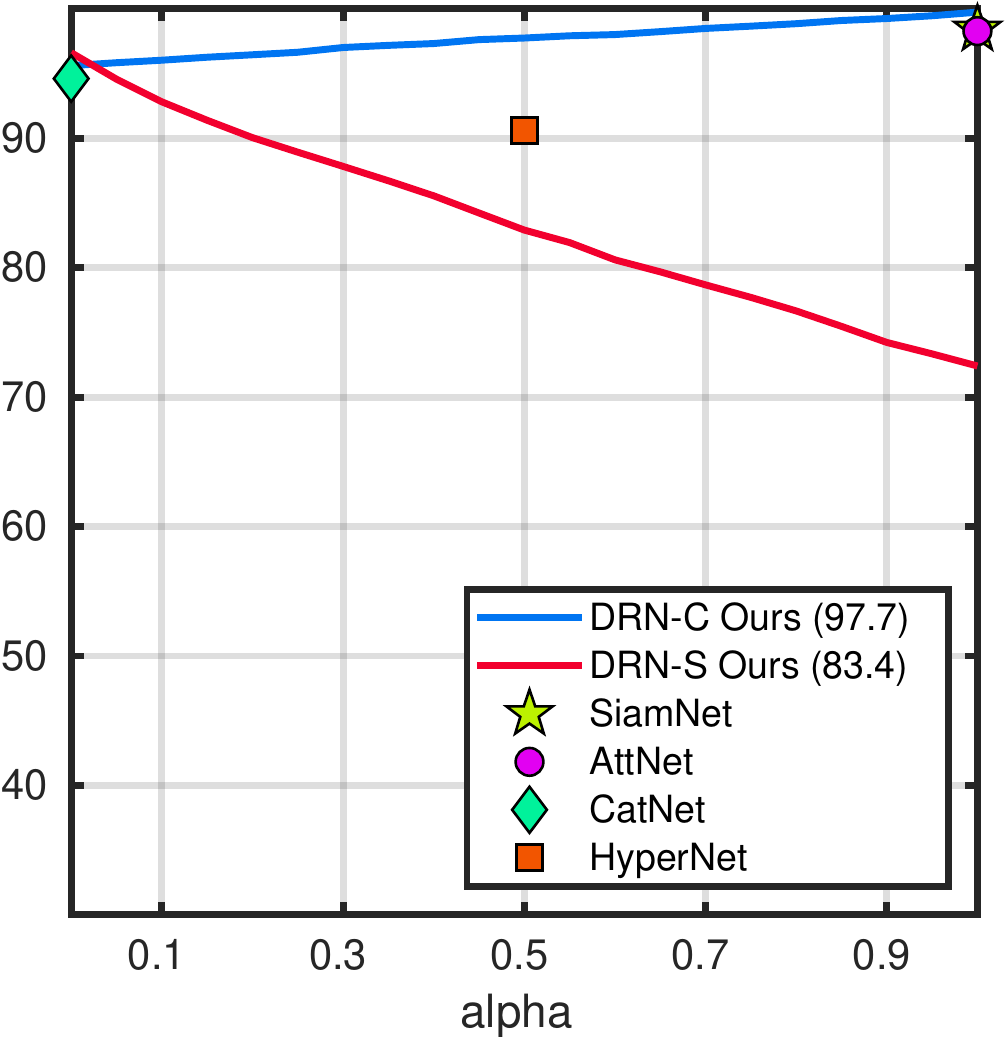}    
    \caption{Top-5}
\end{subfigure}\qquad
\begin{subfigure}[!b]{0.45\linewidth}
    \centering    
    \includegraphics[width=\linewidth]{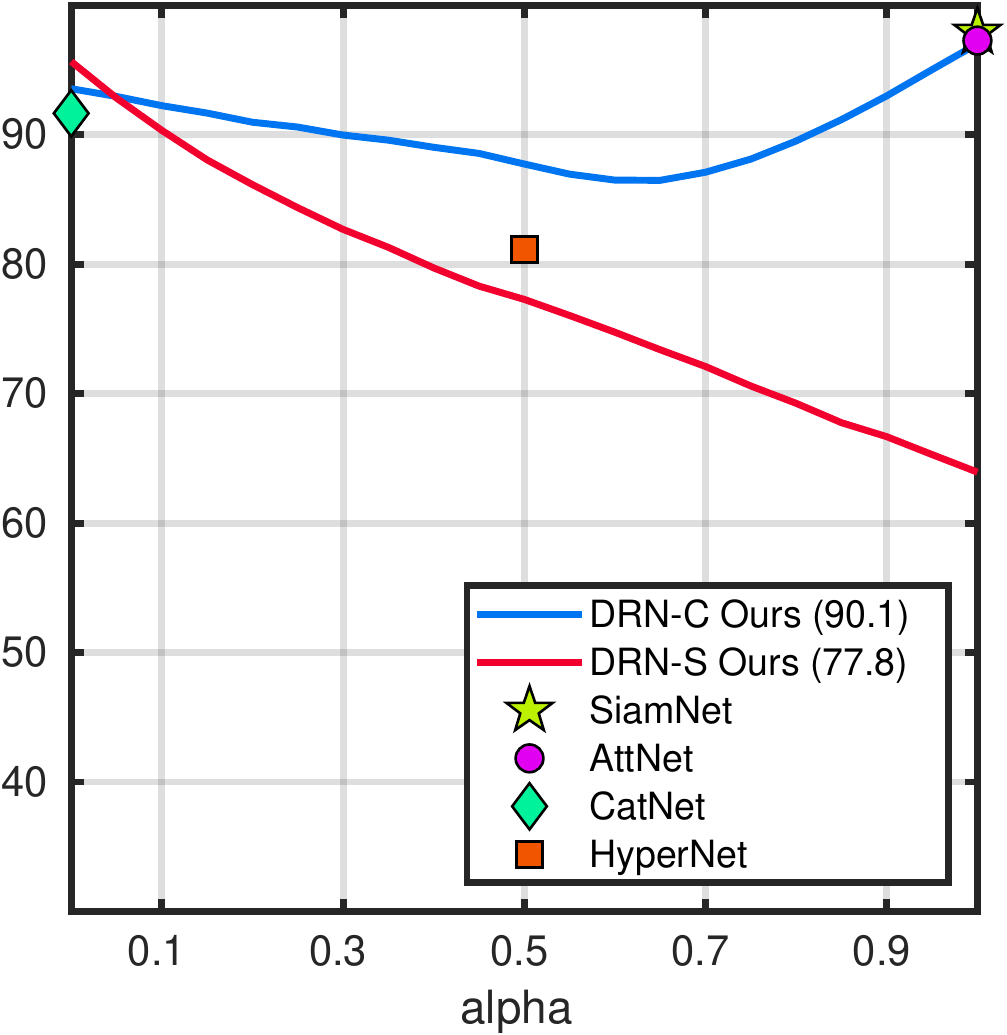}
    \caption{Top-20}
\end{subfigure}
\caption{\small {\bf Performance comparison to baselines.} 
We show {\bf (a)} top-5 and {\bf (b)} top-20 retrieval accuracy of $4$ baseline discrete models (marks) and the two proposed dynamic models (solid lines) as $\alpha$ smoothly changes from $0$ to $1$. 
The proposed models are superior to the baselines and stable across the entire range of $\alpha$, as indicated by the area under the curve shown in parenthesis.}
\label{fig:alpha_acc_baselines}
\vspace{-0.3cm}
\end{figure} 
\paragraph{Comparison to Baselines}
We compare our approach to $4$ well-established \emph{discrete} retrieval models:
\begin{itemize}[label={\textbullet},noitemsep,topsep=1pt, leftmargin=*,itemindent=0pt,labelsep=2pt]
\item Siamese Network (SiamNet~\cite{oh2016deep}): a siamese model that captures the pairwise similarity of samples.
\item Attribute Network (AttNet~\cite{felix2012weak,Huang_2015_ICCV}): a deep model trained for attribute classification.
\item Categorization Network (CatNet~\cite{babenko2014neural,gong2014multiA}): a deep model trained for category classification.
\item Hyper Network (HyperNet \cite{Liu_2017_CVPR}): a deep model trained for the joint objective of predicting both categories and attributes with equal weighting.
\end{itemize}

\noindent
We implement these models and adopt them to our problem and dataset.
All baseline models use the same core CNN structure as our own models (see~\secref{sec:evaluation}).
We represent each image using the output of the last hidden layer of the respective model.
The retrieval is conducted as in the previous experiment, \ie, the query image is compared against all samples in the dataset and ranked using Euclidean distance.

To the best of our knowledge, our model is the first one allowing the retrieval at different operating points of the \retrievalmanifold. 
Since there are no standard performance measures for this task, we introduce a new measure of performance along the continuous \retrievalmanifold, namely the $\alpha$-weighted average of category and attribute accuracy for the top-$K$ retrieved samples:
\begin{equation}
	\mathrm{Top}_K = \alpha\, \mathrm{Top}_K^{A} + (1-\alpha)\,\mathrm{Top}_K^{C}.
\end{equation}
\figref{fig:alpha_acc_baselines} shows the performance of our models compared to the classical retrieval models.
Note that the discrete models generate a \emph{single} static ranking and excels at a specific point on the \retrievalmanifold.
These optimal SROPs of each of these models are highlighted with a marker.
We see that SiamNet and AttNet operate at $\alpha=1$, because their embeddings are optimized with the objective of attribute similarity.
CatNet, on the other hand, operates at $\alpha=0$, since it is optimized with the objective of category classification.
The HyperNet model operates at $\alpha=0.5$, in accordance with the optimization of both category and attribute objectives. 
Our \drn{-S} model is the best-performing model at $\alpha=0$ but declines lower than HyperNet as $\alpha$ goes to $1$. 
We believe this is because the \drn{-S} tries to incorporate attribute information in addition to categorical information, which is substantially harder than a trade-off between both objectives. 
Despite lower performance at certain SROPs, \drn{-S} does exhibit useful behavior in practice.
Finally, our \drn{-C} model shows the best overall performance along the \retrievalmanifold, exhibiting consistently good performance as it traverses between the two extremes of $\alpha=0$ and $\alpha=1$ and even outperforms the discrete models at their optimal SROPs.

\subsection{Traversing Visual Fashion Space}\label{sec:fashion}
We conclude with an application of our dynamic model to the real-world task of fashion exploration, where it allows the interactive traversal between the category and visual style (\ie attributes) of a query fashion item. 

We test our model on two large-scale fashion datasets:
\textbf{1)} The UT-Zappos50K dataset~\cite{yu2014finegrained} which contains around $50,000$ images of footwear items.
The items fall into $4$ main categories (Sandals, Boots, Shoes, and Slippers). 
We select $20$ out of the most frequent tags provided by the vendors as the attribute labels for the images. 
These attributes describe properties such as gender, closure-type, heel-height and toe-style. 
\textbf{2)} The Shopping100K~\cite{Ak_2018_CVPR} which has around $100,000$ images of clothes from $16$ categories (\eg Shirts, Coats and Dresses).
Addiotionally, all images are labeled with $135$ semantic attributes that describe visual properties like collar-type, sleeve-length, fabric and color. 
We split each dataset with a ratio of $1$ to $9$ for testing and training.
We train our \drn{-C} and $3$ discrete baselines that share the same backbone architecture with our model, similar to the setup in our previous experiments.

\begin{figure}[]
\centering
\begin{subfigure}[!b]{0.45\linewidth}
    \centering
    \includegraphics[width=\linewidth]{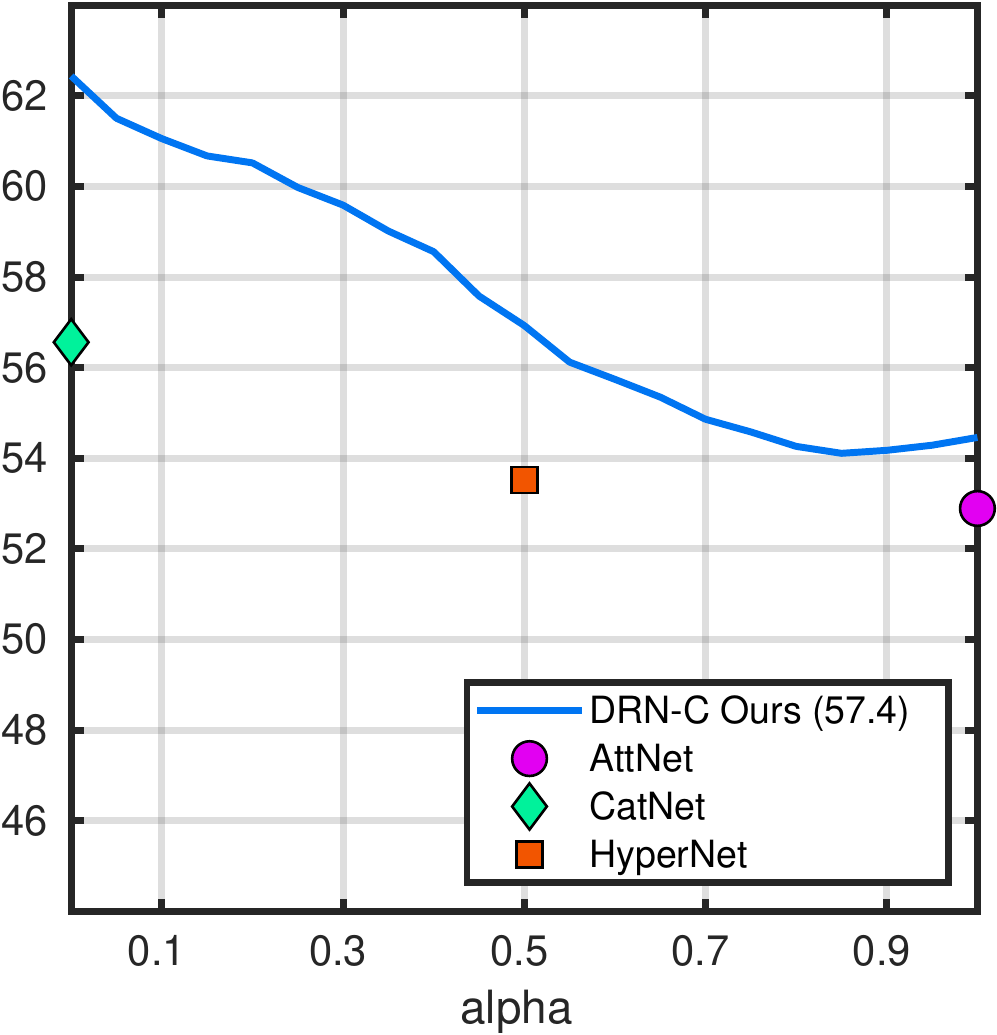}    
    \caption{Shopping100K}
\end{subfigure}\qquad
\begin{subfigure}[!b]{0.45\linewidth}
    \centering    
    \includegraphics[width=\linewidth]{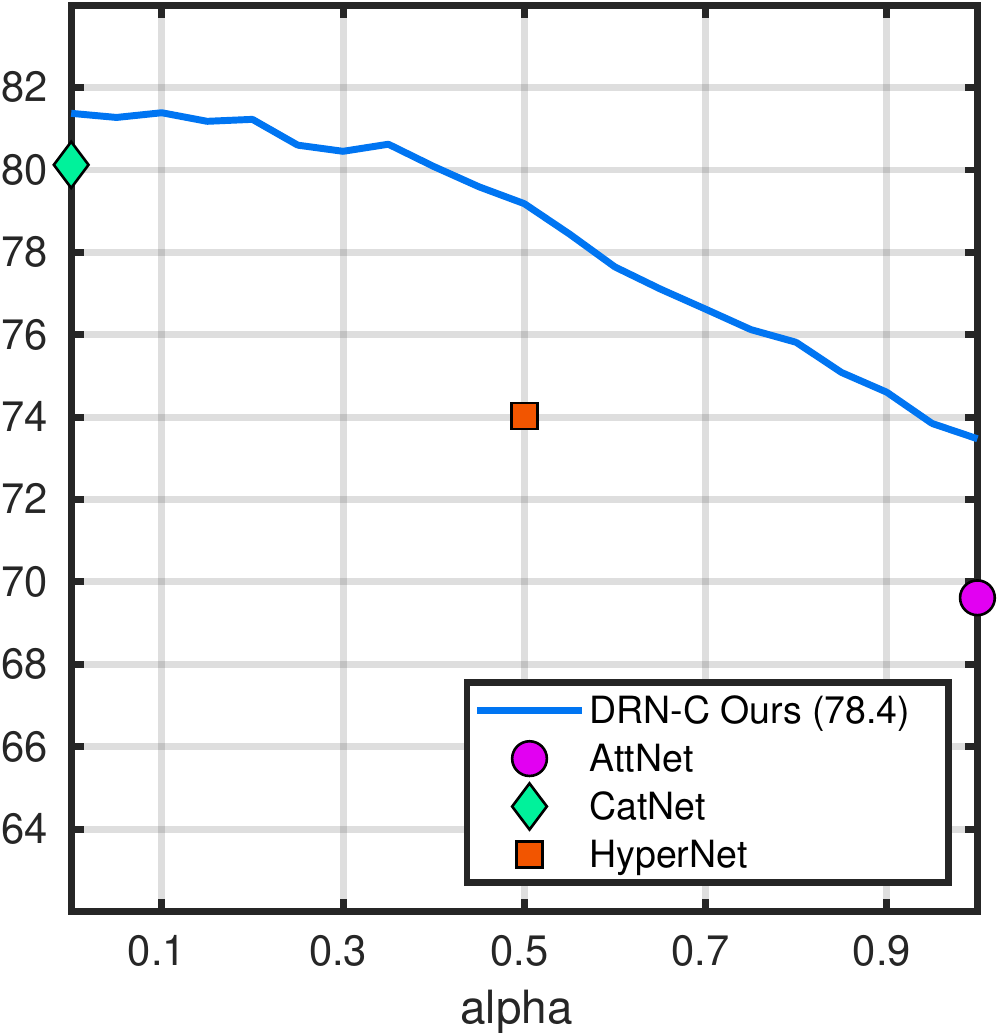}
    \caption{UT-Zappos50K}
\end{subfigure}
\caption{\small {\bf Performance comparison to baselines on fashion retrieval.} 
We show top-20 retrieval accuracy on {\bf (a)} Shopping100K~\cite{Ak_2018_CVPR} and {\bf (b)} UT-Zappos50K~\cite{yu2014finegrained} of $3$ discrete retrieval models (marks) and our proposed model \drn{-C} (solid line) as $\alpha$ smoothly changes from $0$ to $1$.}
\label{fig:alpha_acc_baselines_zappos_shopping}
\vspace{-0.2cm}
\end{figure} 
\begin{figure}[t]
\centering
\begin{subfigure}[!b]{0.325\linewidth}
    \includegraphics[width=1\linewidth]{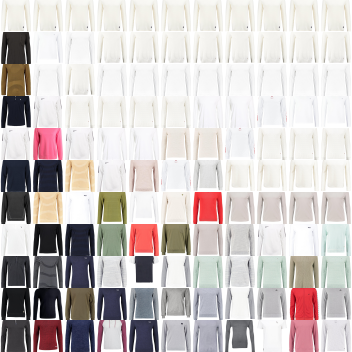}
    \caption{Jumper}
    \label{fig:shopping_qual_a}
\end{subfigure}~~
\begin{subfigure}[!b]{0.325\linewidth}
    \includegraphics[width=1\linewidth]{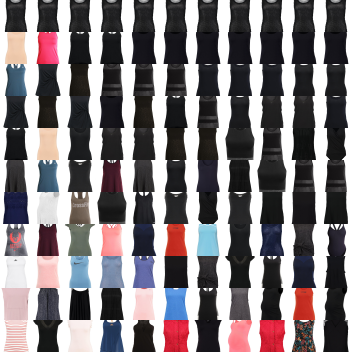}
    \caption{T-Shirt}    
    \label{fig:shopping_qual_b}
\end{subfigure}~~
\begin{subfigure}[!b]{0.325\linewidth}
    \includegraphics[width=1\linewidth]{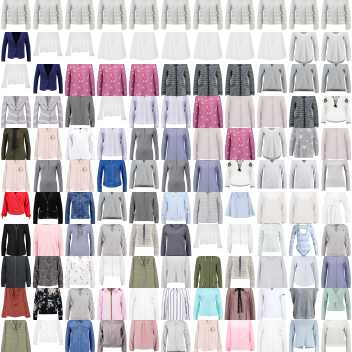}
    \caption{Jacket}
    \label{fig:shopping_qual_c}
\end{subfigure}
\caption{\small {\bf Qualitative results on Shopping100K.} 
As the control parameter $\alpha$ goes from $0$ (left) to $1$ (right), the SROP for each query ($1^{st}$ row) changes from pure category retrieval to style-based retrieval, \ie attribute-based retrieval. Best viewed in color and with zoom-in.}
\label{fig:shopping_qual}
\vspace{-0.6cm}
\end{figure}

\figref{fig:alpha_acc_baselines_zappos_shopping} shows the performance of our model compared to the baselines in terms of top-20 accuracy.
As before, we see that overall our \drn{-C} model outperforms its competitors with a significant margin ($\approx \!4\%$).
Comparing \figref{fig:alpha_acc_baselines_zappos_shopping} to \figref{fig:alpha_acc_baselines}, it is also evident that attribute retrieval on both datasets is substantially more complex than on MNIST Attributes. 
Even AttNet, which should operate optimally at $\alpha=1$, shows a performance closer to HyperNet or worse. 
Upon further inspection, we notice that some of the attributes have strong correlation with category labels. 
For instance, attributes related to heel-heights or toe-styles appear almost exclusively with women's shoes, rendering learning an attribute representation that is disentangled from the associated categories challenging. 
Our model handles these cases well and performs significantly better than all competing methods at their optimal SROPs. 
Similar to \figref{fig:alpha_qual_cls_sim}, we show qualitative examples on Shopping100K in~\figref{fig:shopping_qual}, illustrating our model's ability to smoothly interpolate between categories and styles in fashion space.
  
\section{Conclusion}\label{sec:conclusion}
We introduced the first work to address the retrieval task as a continuous operation traversing the retrieval simplex smoothly between different operating points (SROPs).
We proposed a novel dynamic retrieval model \drn{} that can target the desired SROP along the simplex using a control parameter.
Moreover, we presented key insights on how to optimize training and improve learning of such a model using the proposed dropout decay layer.
We demonstrated the properties and differences between two dynamic learning schemes and highlighted the performance of our model against a set of established discrete deep retrieval models.
Finally, we hope that our findings will be a stepping stone for further research in the direction of traversing the continuous spectrum of image retrieval. 
\balance
{\small
\bibliographystyle{ieee}
\bibliography{mybib}
}

\end{document}